\documentclass[11pt,a4paper]{article}
\usepackage[hyperref]{eacl2021}
\usepackage{times}
\usepackage{float}
\usepackage{amsmath}
\usepackage{graphicx}
\usepackage{latexsym}
\usepackage{algorithm,algpseudocode}

\usepackage{microtype}

\aclfinalcopy 

\title{Have Attention Heads in BERT Learned Constituency Grammar?}

\author{Ziyang Luo \\
  Uppsala University \\
  \texttt{Ziyang.Luo.9588@student.uu.se}}

\date{}

\begin{document}
\maketitle
\begin{abstract}
  With the success of pre-trained language models in recent years, more and more researchers focus on opening the ``black box" of these models. Following this interest, we carry out a qualitative and quantitative analysis of constituency grammar in attention heads of BERT and RoBERTa. We employ the syntactic distance method to extract implicit constituency grammar from the attention weights of each head. Our results show that there exist heads that can induce some grammar types much better than baselines, suggesting that some heads act as a proxy for constituency grammar. We also analyze how attention heads' constituency grammar inducing (CGI) ability changes after fine-tuning with two kinds of tasks, including sentence meaning similarity (SMS) tasks and natural language inference (NLI) tasks. Our results suggest that SMS tasks decrease the average CGI ability of upper layers, while NLI tasks increase it. Lastly, we investigate the connections between CGI ability and natural language understanding ability on QQP and MNLI tasks.
\end{abstract}



\section{Introduction}

Recently, pre-trained language models have achieved great success in many natural language processing tasks \citep{devlin-etal-2019-bert,NIPS2019_8812}, including sentiment analysis \citep{liu2019roberta},  question answering \citep{Lan2020ALBERT} and constituency parsing \citep{ijcai2020-560}, to name a few. Though these models have become more and more popular in many NLP tasks, they are still ``black boxes". What they have learned, and why and when they perform well remain unknown. To open these ``black boxes", researchers have used many methods to analyze the linguistic knowledge that these models encode \citep{goldberg2019assessing,clark-etal-2019-bert,hewitt-manning-2019-structural,kim2020pretrained}.

Pre-trained language models use self-attention mechanism in each layer to compute the internal representations of each token. In this work, we investigate the hypothesis that some attention heads in pre-trained language models have learned constituency grammar. We use an unsupervised constituency parsing method to extract constituency trees from each attention heads of BERT \citep{devlin-etal-2019-bert} and RoBERTa \citep{liu2019roberta} before and after fine-tuning. This method computes the syntactic distance between every two adjacent words and generates a constituency parsing tree recursively. We analyze the extracted constituency parsing trees to investigate whether specific attention heads induce constituency grammar better than baselines, and which types of constituency grammars they learn best. 

In prior work, \citet{kim2020pretrained} show that some layers of pre-trained language models exhibit syntactic structure akin to constituency grammar to some degree. However, they do not analyze how fine-tuning affects models. We first follow their methods to extract constituency grammar from BERT and RoBERTa. Then, we use the same approach to analyze BERT and RoBERTa after fine-tuning. To the best of our knowledge, we are the first to investigate how fine-tuning affects the constituency grammar inducing (CGI) ability of attention heads. We fine-tune them on two types of GLUE natural language understanding (NLU) tasks \citep{N18-1101,wang-etal-2018-glue}. The first type is the sentence meaning similarity (SMS) task. We fine-tune our models on two datasets, QQP \footnote{https://www.quora.com/q/quoradata/First-Quora-Dataset-Release-Question-Pairs} and STS-B \citep{cer-etal-2017-semeval}. The second type is the natural language inference (NLI) task. We fine-tune our models on two datasets, MNLI \citep{N18-1101} and QNLI \citep{rajpurkar-etal-2016-squad,wang-etal-2018-glue}.  Lastly, we investigate the relations between CGI ability of attention heads and natural language understanding ability on QQP and MNLI tasks.

The findings of our study are as follows:
\begin{enumerate}
    \item Attention heads in the higher layers of BERT and the middle layers of RoBERTa have better constituency grammar inducing (CGI) ability. Some heads act as a proxy for some constituency grammar types, but all heads do not appear to fully learn constituency grammar.
    \item The sentence meaning similarity task decreases the average CGI ability in the higher layers. The natural language inference task increases it in the higher layers.
    \item For QQP and MNLI tasks, attention heads with better CGI ability are more important for BERT. However, this relation is different in RoBERTa.
\end{enumerate}

\section{Related Work}

Many works have proposed methods to induce constituency grammar and extract constituency trees from the attention heads of the transformer-based model. \citet{marecek-rosa-2018-extracting} aggregate all the attention distributions through the layers and get an attention weight matrix. They extract binary constituency tree and undirected dependency tree from this matrix. \citet{kim2020pretrained} use the attention distribution and internal vector representation to compute Syntactic Distance \citep{shen-etal-2018-straight} between every two adjacent words to draw constituency trees from raw sentences without any training.

Additionally, researchers have investigated how fine-tuning affects syntactic knowledge that BERT learns. \citet{kovaleva-etal-2019-revealing} use the subset of GLUE tasks \citep{wang-etal-2018-glue} to fine-tune BERT-base model. They find that fine-tuning does not change the self-attention patterns. They also find that after fine-tuning, the last two layers' attention heads undergo the largest changes. \citet{htut2019attention} investigates whether fine-tuning affects the dependency syntax in BERT attentions. They find that fine-tuning does not have great effects on attention heads' dependency syntax inducing ability. \citet{zhao-bethard-2020-berts} investigate the negation scope linguistic knowledge in BERT and RoBERTa's attention heads before and after fine-tuning. They find that after fine-tuning, the average attention heads are more sensitive to negation.

While there are some prior works analyzing attention heads in BERT, we believe we are the first to analyze the constituency grammar learned by fine-tuned BERT and RoBERTa models.

\section{Methods}

\subsection{Transformer and BERT}

Transformer \citep{NIPS2017_7181} is a neural network model based on self-attention mechanism. It contains multiple layers and each layer contains multiple attention heads. Each attention head takes a sequence of input vectors $h=[h_1,...,h_n]$ corresponding to the n tokens. An attention head will transform each vector $h_i$ into query $q_i$, key $k_i$, and value $v_i$ vectors. Then it computes the output $o_i$ by a weighted sum of the value vectors.

\begin{equation}
    a_{ij}=\frac{\exp(q_i^Tk_j)}{\sum_{t=1}^n\exp(q_i^Tk_t)}
\end{equation}
\begin{equation}
    o_i=\sum_{j=1}^na_{ij}v_j
\end{equation}
Attention weights distribution of each token can be viewed as the ``importance" from other tokens in the sentence to the current token.

BERT is a Transformer-based pre-trained language model. It is pre-trained on BooksCorpus \citep{zhu2015aligning} and English Wikipedia with masked language model (MLM) objective and next sentence prediction (NSP) objective. RoBERTa is a modified version of BERT. It removes the NSP pre-training objective and training with much larger mini-batches and learning rates. We use the uncased base size of BERT and base size of RoBERTa which have 12 layers and each layer contains 12 attention heads. Our models are downloaded from Hugging Face's Transformers Library \footnote{https://huggingface.co/models} \citep{wolf2020huggingfaces}.

\subsection{Analysis Methods}

We aim to analyze constituency grammar in attention heads. We use a method to extract constituency parsing trees from attention distributions. This method operates on the attention weight matrix $W\in (0, 1)^{T\times T}$ for every head at a given layer, where T is the number of tokens in the sentence.\\

\noindent \textbf{Method: Syntactic Distance to Constituency Tree} To extract complete valid constituency parsing trees from the attention weights for a given layer and head, we follow the method of \citet{kim2020pretrained} and treat every row of the attention weight matrix as attention distribution of each token in the sentence. As in \citet{kim2020pretrained}, we compute the syntactic distance vector \textbf{d}$=[d_1,d_2,...,d_{n-1}]$ for a given sentence $w_1,...,w_n$, where $d_i$ is the syntactic distance between $w_i$ and $w_{i+1}$. Each $d_i$ is defined as follows:
\begin{equation}
    d_i=f(g(w_i), g(w_{i+1})),
\end{equation}
where $f(\cdot,\cdot)$ and $g(\cdot)$ are a distance measure function and feature extractor function. We use Jensen-Shannon function to measure the distance between each attention distribution. Appendix \ref{JSD} gives a brief introduction of this function. $g(w_i)$ is equal to the $i^{th}$ row of the attention matrix $W$.

To introduce the right-skewness bias for English constituency trees, we follow \citet{kim2020pretrained} by adding a linear bias term to every $d_i$:
\begin{equation}
    \hat{d}_i=d_i+\lambda \cdot Mean(\textbf{d})\times \left(1-\frac{i-1}{m-1}\right),
\end{equation}
where $m=n-1$ and $\lambda$ is set to $1.5$.

After computing the syntactic distance, we use the algorithm introduced by \citet{shen-etal-2018-straight} to get the target constituency tree. Appendix \ref{alg} describes this algorithm.

Constituency parsing is a word-level task, but BERT uses byte-pair tokenization \citep{sennrich-etal-2016-neural}. This means that some words are tokenized into subword units. Therefore, we need to convert token-to-token attention matrix to word-to-word attention matrix. We merge the non-matching subword units and compute the means of the attention distributions for the corresponding rows and columns. We use two baselines in our experiments. They are left-branching and right-branching trees.

\subsection{Experiments Setup}
In our experiments, we use an unsupervised constituency parsing method to induce constituency grammar on WSJ Penn Treebank (PTB, \citet{marcus-etal-1993-building}) without any training. We use the standard split of the dataset-23 for testing. We use sentence-level F1 (S-F1) score to evaluate our models. In addition, we also report label recall scores for six main phrase categories: SBAR, NP, VP, PP, ADJP, and ADVP.

\section{Results and Analysis}

\subsection{Constituency Grammar in Attention Heads before Fine-tuning}

\begin{figure}
    \centering
    \includegraphics[height=5cm]{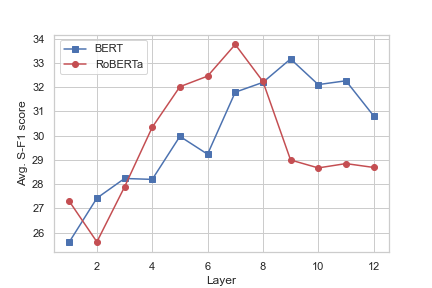}
    \caption{Average constituency parsing S-F1 score of each layer in BERT and RoBERTa.}
    \label{fig:bert-roberta}
\end{figure}

In this part, our goal is to understand how constituency grammar is captured by different attention heads in BERT and RoBERTa before fine-tuning. First, we investigate the common patterns of attention heads' constituency grammar inducing (CGI) ability in BERT and RoBERTa. From Figure \ref{fig:bert-roberta}, we can find that the CGI ability of the higher layers of BERT is better than the lower layers. However, the middle layers of RoBERTa are better than the other layers. In appendix \ref{heatmap}, two heatmaps of every heads' S-F1 score in BERT and RoBERTa also show such patterns.

\begin{table*}
  \begin{center}
    \begin{tabular}{lccccccc}
    \hline
    Models & S-F1 & SBAR & NP & VP & PP & ADJP & ADVP\\
    \hline
    \hline
    \textbf{Baselines} & & & & & & &\\
    Left-branching Trees  & 8.73 & 5.46\% & 11.33\% & 0.82\% & 5.02\% & 2.46\% & 8.04\%\\
    Right-branching Trees & 39.46 & 68.76\% & 24.89\% & 71.76\% & 42.43\% & 27.65\% & 38.11\%\\
    \hline
    \hline
    \textbf{Pre-trained LMs} & & & & & & & \\
    BERT & 39.47 & 67.32\% & 46.48\% & 68.82\% & 57.26\% & 46.39\% & 65.03\%\\
    BERT-QQP & \textcolor{red}{39.97} & 67.32\% & \textcolor{blue}{45.39}\% & \textcolor{blue}{68.79}\% & \textcolor{blue}{50.71}\% & \textcolor{blue}{45.01}\% & \textcolor{blue}{61.54}\%\\
    BERT-STS-B & \textcolor{red}{39.48} & 67.32\% & \textcolor{blue}{44.16}\% & 68.82\% & \textcolor{blue}{56.68}\% & \textcolor{red}{48.39}\% & \textcolor{blue}{57.69}\%\\
    BERT-QNLI & \textcolor{red}{39.74} & 67.32\% & \textcolor{red}{50.96}\% & \textcolor{blue}{68.81}\% & \textcolor{red}{65.38}\% & \textcolor{blue}{46.08}\% & \textcolor{blue}{63.29}\%\\
    BERT-MNLI & \textcolor{red}{39.66} & 67.32\% & \textcolor{red}{44.89}\% & \textcolor{blue}{68.75}\% & \textcolor{red}{62.81}\% & \textcolor{red}{49.16}\% & \textcolor{blue}{64.69}\%\\
    \hline
    RoBERTa & 39.60 & 67.43\% & 47.92\% & 69.35\% & 56.53\% & 49.00\% & 66.43\%\\
    RoBERTa-QQP & \textcolor{blue}{39.41} & \textcolor{blue}{66.70}\% & \textcolor{blue}{43.02}\% & \textcolor{red}{69.45}\% & \textcolor{blue}{51.06}\% & \textcolor{blue}{43.16}\% & \textcolor{blue}{60.84}\%\\
    RoBERTa-STS-B & \textcolor{red}{40.36} & \textcolor{blue}{66.76}\% & \textcolor{blue}{46.82}\% & \textcolor{red}{69.50}\% & \textcolor{blue}{54.91}\% & \textcolor{blue}{46.54}\% & \textcolor{blue}{64.34}\%\\
    RoBERTa-QNLI & \textcolor{red}{43.95} & \textcolor{blue}{66.76}\% & \textcolor{red}{52.51}\% & \textcolor{red}{69.48}\% & \textcolor{red}{58.30}\% & \textcolor{blue}{48.39}\% & \textcolor{red}{69.23}\%\\
    RoBERTa-MNLI & \textcolor{red}{40.41} & \textcolor{blue}{66.76}\% & \textcolor{red}{47.97}\% & \textcolor{red}{69.42}\% & \textcolor{red}{57.50}\% & \textcolor{blue}{47.77}\% & \textcolor{red}{68.88}\%\\
    \hline
    \end{tabular}
    \caption{Highest constituency parsing scores of all models. \textcolor{blue}{Blue} score means that this score is lower after fine-tuning. \textcolor{red}{Red} score means that this score is higher after fine-tuning.}
    \label{tab:results}
  \end{center}
\end{table*}

Table \ref{tab:results} describes the S-F1 scores of the best attention heads of BERT and RoBERTa. We also choose the best recall for each phrase type. We observe that the S-F1 scores of BERT and RoBERTa are only slightly better than the right-branching baseline. This implies that the attention heads in BERT and RoBERTa do not appear to fully learn constituency grammar. However, they outperform the baselines by a large margin for noun phrase (NP), preposition phrase (PP), adjective phrase (ADJP), and adverb phrase (ADVP). This implies that the attention heads in BERT and RoBERTa only learn a part of constituency grammar.

\subsection{Constituency Grammar in Attention Heads after Fine-tuning}

\begin{figure}
    \centering
    \includegraphics[height=5cm]{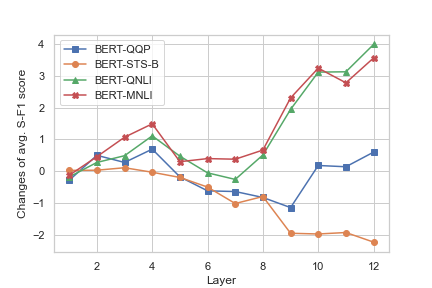}
    \caption{Changes of average S-F1 score of each layer in BERT after fine-tuning.}
    \label{fig:bertchange}
\end{figure}

\begin{figure}
    \centering
    \includegraphics[height=5cm]{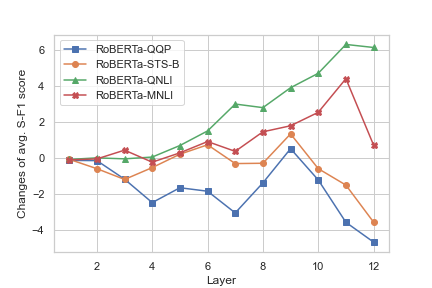}
    \caption{Changes of average S-F1 score of each layer in RoBERTa after fine-tuning.}
    \label{fig:robertachange}
\end{figure}

In this part, we fine-tune BERT and RoBERTa with four downstream tasks, QQP, STS-B, QNLI, and MNLI. These four tasks can be divided into two types. The first type is the sentence meaning similarity task (SMS), including QQP and STS-B. This task requires models to determine whether two sentences have the same meaning. The second type is the natural language inference task (NLI), including QNLI and MNLI. This task requires models to determine whether the first sentence can infer the second sentence. We want to analyze how these two kinds of downstream tasks affect constituency grammar inducing (CGI) ability of attention heads in BERT and RoBERTa.

Figure \ref{fig:bertchange} and Figure \ref{fig:robertachange} show that these four tasks do not have much influence on BERT and RoBERTa for the lower layers. For the higher layers, fine-tuning with NLI tasks can increase the average CGI ability of attention heads in BERT and RoBERTa. However, fine-tuning with SMS tasks harms it. 

Table \ref{tab:results} shows that fine-tuning can increase the highest constituency parsing scores of all models except RoBERTa-QQP. However, fine-tuning with SMS tasks decreases the ability of attention heads to induce NP, PP, ADJP, and ADVP. For BERT, NLI tasks can increase the ability of attention heads to induce NP, PP. For RoBERTa, NLI tasks can increase the ability of attention heads to induce NP, VP, PP, and ADVP.


\subsection{Constituency Grammar Inducing Ability and Natural Language Understanding Ability}\label{sec:NLU}

In this part, we analyze the relations between constituency grammar inducing (CGI) ability and natural language understanding (NLU) ability on QQP and MNLI tasks. We use the performance of BERT and RoBERTa to evaluate their NLU ability. We report the scores on the validation, rather than test data, so the results are different from the original BERT paper.

First, we sort all attention heads in each layer based on their S-F1 scores before fine-tuning. Then we use the method in \citet{NIPS2019_9551} to mask the top-k/bottom-k ($k=1,...,11$) attention heads in each layer and compute the accuracy on two downstream tasks, QQP and MNLI.

\begin{figure}
    \centering
    \includegraphics[height=5cm]{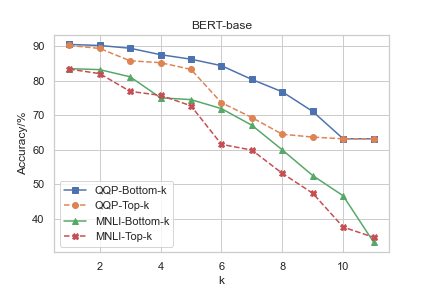}
    \caption{QQP dev and MNLI dev-matched accuracy after masking the top-k/bottom-k attention heads in each layer of BERT-QQP and BERT-MNLI.}
    \label{fig:BERT-Acc}
\end{figure}

\begin{figure}
    \centering
    \includegraphics[height=5cm]{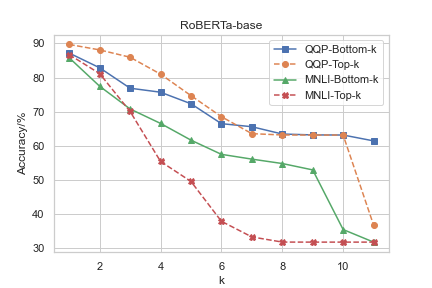}
    \caption{QQP dev and MNLI dev-matched accuracy after masking the top-k/bottom-k attention heads in each layer of RoBERTa-QQP and RoBERTa-MNLI.}
    \label{fig:RoBERTa-Acc}
\end{figure}

Figure \ref{fig:BERT-Acc} shows that downstream tasks accuracy scores decrease quicker when we have masked the top-k attention heads in BERT. Especially for the QQP task, after masking the bottom-7 attention heads in all layers, accuracy is still higher than 80\%, which is more than 10\% higher than masking the top-7 attention heads.

Figure \ref{fig:RoBERTa-Acc} shows that masking RoBERTa has different results from BERT. For the QQP task, when k is smaller or equal to 6, masking the bottom-k attention heads in all layers decreases faster. For the MNLI task, when k is 1 or 2, masking the bottom-k heads decreases also faster. When k is larger than 6 in the QQP task and 2 in the MNLI task, masking the top-k heads decreases faster.

For BERT, the results show that attention heads with better CGI ability are more important for a model to gain NLU ability on these two tasks. For RoBERTa, the connections between CGI ability and NLU ability are not as strong as BERT. For the MNLI task, we still can find that better CGI ability is more important for NLU ability. However, better heads are not so important for QQP task.

\section{Discussion}

The experiments detailed in the previous sections point out that the attention heads in BERT and RoBERTa does not fully learn much constituency grammar knowledge. Even after fine-tuning with downstream tasks, the best constituency parsing score does not change much. Our results are similar to \citet{htut2019attention}. They also point out that the attention heads do not fully learn much dependency syntax. Fine-tuning does not affect these results. This raises an interesting question: do attention heads not contain syntax (constituency or dependency) information? If this is true, where does BERT encode this information? Also, is syntax information not important for BERT to understand language? Our simple experiment in §\ref{sec:NLU} shows that the attention heads with better constituency grammar inducing ability are not important for RoBERTa on QQP task. \citet{glavas2020supervised} also point out that leveraging explicit formalized syntactic structures provides zero to negligible impact on NLU tasks. The relations between syntax and BERT's NLU ability still need to be further analyzed.

\section{Conclusion}

In this work, we investigate whether the attention heads in BERT and RoBERTa have learned constituency grammar before and after fine-tuning. We use a method to extract constituency parsing trees without any training, and observe that the upper layers of BERT and the middle layers of RoBERTa show better constituency grammar ability. Certain attention heads better induce specific phrase types, but none of the heads show strong constituency grammar inducing (CGI) ability. Furthermore, we observe that fine-tuning with SMS tasks decreases the average CGI ability of upper layers, but NLI tasks can increase it. Lastly, we mask some heads based on their parsing S-F1 scores. We show that attention heads with better CGI ability are more important for BERT on QQP and MNLI tasks. For RoBERTa, better heads are not so important on QQP task.

One of the directions for future research would be to further study the relations between downstream tasks and the CGI ability in attention heads and to explain why different tasks have different effects.

\section*{Acknowledgments}
This project grew out of a master course project for the Fall 2020 Uppsala University 5LN714, \textit{Language Technology: Research and Development}. We would like to thank Sara Stymne and Ali Basirat for some great suggestions and the anonymous reviewers for their excellent feedback.

\bibliography{eacl2021}
\bibliographystyle{acl_natbib}

\appendix
\section{Jensen-Shannon Distance Measure Function}\label{JSD}

Jensen-Shannon function measures the distance between two distributions. Suppose that we have two distributions $P$ and $Q$, the Jensen-Shannon Distance is defined as
\begin{equation}
    JSD(P||Q)=\left(\frac{D_{KL}(P||M)+D_{KL}(Q||M)}{2}\right)^{\frac{1}{2}},
\end{equation}
where $M=(P+Q)/2$ and $D_{KL}(A||B)=\sum_wA(w)\log(A(w)/B(w))$.

\section{Syntactic Distances to Constituency Trees Algorithm}\label{alg}

\begin{algorithm}[H]
\caption{Syntactic Distances to Constituency Trees Algorithm \citep{shen-etal-2018-straight}}\label{euclid}
\begin{algorithmic}[1]
\State $S=[w_1,w_2,...,w_n]:$ a sentence with n words.
\State $\textbf{d}=[d_1,d_2,...,d_{n-1}]:$ a sequence of distances between every two adjacent words.
\Function{Tree}{S, \textbf{d}}
\If {\textbf{d} is empty}
\State $\textit{node} \gets \text{Leaf}(S[0])$
\Else
\State $\textit{i} \gets \text{arg}\max_i(\textbf{d})$
\State $\text{lchild} \gets \text{TREE}(S_{\leq i}, \textbf{d}_{<i})$
\State $\text{rchild} \gets \text{TREE}(S_{> i}, \textbf{d}_{>i})$
\State $\textit{node} \gets \text{Node(lchild, rchild)}$
\EndIf
\Return \textit{node}
\EndFunction
\end{algorithmic}
\end{algorithm}

\section{BERT and RoBERTa Heatmaps}\label{heatmap}

In this section, we present two heatmaps of S-F1 score of each heads in BERT and RoBERTa. Row represents layer and column represents head.

\begin{figure}[H]
    \centering
    \includegraphics[height=6cm]{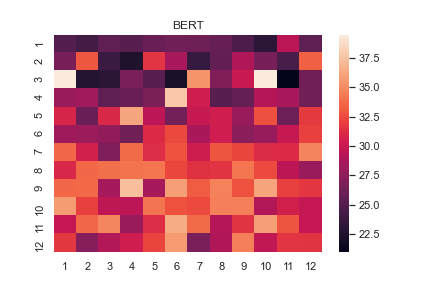}
    \caption{S-F1 score of each heads in BERT.}
    \label{fig:bertheat}
\end{figure}
\begin{figure}[H]
    \centering
    \includegraphics[height=6cm]{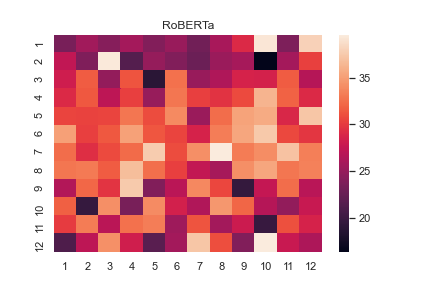}
    \caption{S-F1 score of each heads in RoBERTa.}
    \label{fig:robertheat}
\end{figure}

\end{document}